\documentclass[a4paper,conference]{IEEEtran}

\usepackage{makecell}
\usepackage{graphicx}
\usepackage[colorlinks,linkcolor=black,citecolor=black]{hyperref}
\usepackage{multirow}
\IEEEoverridecommandlockouts

\ifCLASSINFOpdf

\else

\fi

\begin{document}

\title{Feature Aggregation in Zero-Shot Cross-Lingual Transfer Using Multilingual BERT}

\author{\IEEEauthorblockN{Beiduo Chen and Wu Guo\thanks{Thanks to the National Natural Science Foundation of China (Grant No. U1836219) for funding.}}
\IEEEauthorblockA{National Engineering Laboratory for\\ Speech and Language Information Processing\\
University of Science and Technology of China\\
Hefei, China\\
Email: bdchen@mail.ustc.edu.cn,\\guowu@ustc.edu.cn}

\and
\IEEEauthorblockN{Quan Liu and Kun Tao}
\IEEEauthorblockA{State Key Laboratory of Cognitive Intelligence\\
iFLYTEK Research\\
Hefei, China\\
Email: \{quanliu,kuntao3\}@iflytek.com}}

\maketitle

\begin{abstract}
Multilingual BERT (mBERT), a language model pre-trained on large multilingual corpora, has impressive zero-shot cross-lingual transfer capabilities and performs surprisingly well on zero-shot POS tagging and Named Entity Recognition (NER), as well as on cross-lingual model transfer. 
At present, the mainstream methods to solve the cross-lingual downstream tasks are always using the last transformer layer’s output of mBERT as the representation of linguistic information. 
In this work, we explore the complementary property of lower layers to the last transformer layer of mBERT. 
A feature aggregation module based on an attention mechanism is proposed to fuse the information contained in different layers of mBERT. 
The experiments are conducted on four zero-shot cross-lingual transfer datasets, and the proposed method obtains performance improvements on key multilingual benchmark tasks XNLI (+1.5 \%), PAWS-X (+2.4 \%), NER (+1.2 F1), and POS (+1.5 F1). 
Through the analysis of the experimental results, we prove that the layers before the last layer of mBERT can provide extra useful information for cross-lingual downstream tasks and explore the interpretability of mBERT empirically.
\end{abstract}

\IEEEpeerreviewmaketitle

\section{Introduction}\label{sect1}
Zero-shot cross-lingual transfer aims at building models for a target language by reusing knowledge acquired from a source language \cite{DBLP:journals/jair/RuderVS19}. 
Multilingual representation spaces aim to capture meaning across language boundaries and that way (at least conceptually) enables the cross-lingual transfer of task-specific NLP models from resource-rich languages with large annotated datasets to resource-lean languages without (m)any labeled instances \cite{DBLP:conf/acl/JoshiSBBC20}. 
Multilingual BERT (mBERT) \cite{DBLP:conf/naacl/DevlinCLT19} is the mainstream deep neural language model pre-trained on the concatenation of monolingual Wikipedia corpora from 104 languages, and it performs surprisingly well on zero-shot POS tagging and Named Entity Recognition (NER), as well as on cross-lingual model transfer \cite{DBLP:conf/acl/PiresSG19}.


\begin{figure}

\includegraphics[width=0.5\textwidth]{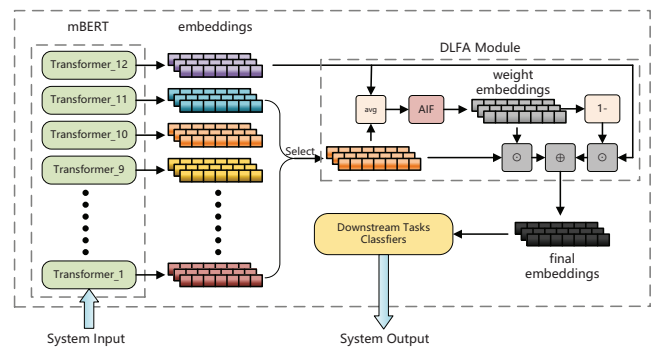}
\caption{The overall structure of the proposed system. }  \label{fig1}

\end{figure}

Due to the strong ability of mBERT, many downstream tasks for zero-shot cross-lingual transfer set mBERT as the pre-trained language model. 
There are two common ways to realize downstream tasks’ targets.
One way is to feed the task-specifically processed data to the pre-trained mBERT.
The other way is to append an extra adapting network to the original mBERT model.
Both methods need to fine-tune the mBERT on source language data, or generated data from the source language. 
To further improve the performance, Qin et al. proposed a data augmentation framework for aligning representations from source and multiple target languages \cite{DBLP:conf/ijcai/QinN0C20}. 
Chung et al. described a novel clustering-based multilingual vocabulary generation algorithm to balance the trade-off between optimizing for cross-lingual subword sharing and the need for robust representation of individual languages \cite{DBLP:conf/emnlp/ChungGTR20}. 
Vidoni et al. introduced orthogonal language and task adapters to add extra trainable parameters that enabled efficient fine-tuning of pre-trained transformers \cite{DBLP:journals/corr/abs-2012-06460}. 
Tasks like NER were usually solved by adding CRF layers behind mBERT \cite{DBLP:journals/eswa/ArasMDC21,DBLP:journals/corr/abs-2203-03216}.

In general, the outputs of the last transformer layer of mBERT are always regarded as the embeddings representing all the information extracted from the language model. 
In this paper, we explore the information of lower layers of mBERT and find it complementary to the last one. 
There are amount of articles focusing on this area at present. 
Jawahar et al. find that each layer of BERT encodes different structural information of language and different downstream tasks demand inconsistent linguistic information \cite{DBLP:conf/acl/JawaharSS19}. 
de Vries et al. analyze which layer has more effects on specific downstream tasks \cite{DBLP:conf/emnlp/VriesCN20}. 
Chen et al. use the first layer of mBERT to help construct the sentence embedding \cite{DBLP:journals/corr/abs-2202-13083}. 
Muller et al. find that mBERT seems like a traditional two-step standard cross-lingual pipeline \cite{DBLP:journals/jair/RuderVS19}: a multilingual encoder in the lower layers, which aligns hidden representations across languages and is critical for cross-language transfer; and a task-specific, language-agnostic predictor that has little effect on cross-language transfer, in the upper layers \cite{DBLP:conf/eacl/MullerESS21}. 
Inspired by these articles, we take full advantage of the information in lower layers of mBERT, and selectively integrate them with the information of the last layer for different downstream tasks to promote the model’s ability on cross-lingual transfer as well as learning language structure.

Since lower layers have different effects on specific downstream tasks, there is no need to integrate the information of all lower layers. 
We selectively aggregate the output features of one lower layer with the last transformer layer for computing efficiency. 
And the double layers feature aggregation (DLFA) module based on an attention mechanism \cite{DBLP:conf/nips/VaswaniSPUJGKP17,DBLP:journals/pami/HuSASW20} is proposed to select weight and fuse the features in an element-wise way. 
The final output embeddings are fed into the backend classifiers for different downstream tasks in the XTREME benchmark \cite{DBLP:conf/icml/HuRSNFJ20}. 
After analyzing a host of results, we come to a consistent conclusion on mBERT that lower layers provide more cross-lingual information while upper layers provide more language structure information.

In all, we accomplish this paper with the following contributions:
\begin{itemize}
\item[-] 
We prove that the output of layers before the last layer can provide supplementary information to the last layer of mBERT for different zero-shot cross-lingual downstream tasks. The optimal dynamic equilibrium between cross-lingual capability and language-structured ability of mBERT is discussed.

\item[-]
We design a feature aggregation module based on an attention mechanism to fuse information from two transformer layers.

\item[-]
Experimental results on four cross-lingual downstream datasets show that our method improves the performance of mBERT on all tasks compared to the baseline and is generalized in various situations.

\end{itemize}

\section{Method}\label{sect2}
In this section, we’re going to introduce the feature aggregation module based on an attention mechanism. 
The double layers feature aggregation (DLFA) module containing the attentional information fusion (AIF) module is proposed. 
The overall structure of our system is shown in Fig.~\ref{fig1}. 
Conventional methods only adopt the last layer embedding of mBERT for downstream tasks. 
However, we use the DLFA module to integrate the representations from the last and from one of the lower layers, and then the fusion embeddings are fed into the classifiers for downstream tasks.

\subsection{Attentional Information Fusion}

As is known, the representation generated from transformers in mBERT is a tensor with the dimension as \( B\times T\times E \), in which \( B \)  is the batch size, \( T \) represents the number of tokens in one sentence (namely the sentence length), and \( E \) means the dimension of the word embedding. 
To aggregate features from different layers of mBERT effectively, we expect the AIF could obtain information dynamically according to the requirements of different downstream tasks.

\begin{figure}
\centering
\includegraphics[width=0.48\textwidth]{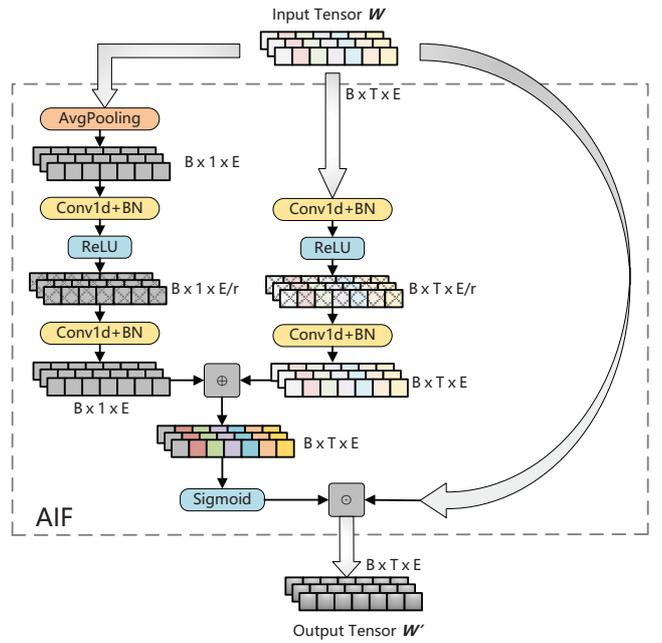}
\caption{The architecture of the proposed attentional information fusion (AIF) module. The AIF extracts global and local information via two branches and element-wisely multiplies the result with the input tensor.} \label{fig2}
\centering
\end{figure}

The earliest idea of the AIF's structure came from SENet \cite{DBLP:journals/pami/HuSASW20}, which expanded and compressed the dimension of features in hope of getting a more efficient and focused representation. 
We introduce this idea to one-dimensional text features, and thus there are two convolution layers in the AIF. 
As Fig.~\ref{fig2} shows, we set the module’s input as \( {W}\in{R}^{B\times T\times E} \). There are two contextual aggregation branches in the module. 
The left one aims to extract global information of input sentences, and the generated weight \( {W}_{{Global}} \) is as follows:

\begin{equation}
{W}_{{Global}}=\mathcal{B}({C}_{G2}(ReLU(\mathcal{B}({C}_{G1}(Pool({W}))))))
\end{equation}

\noindent where \( \mathcal{B}(\cdot) \) denotes the Batch Normalization (BN) operation \cite{DBLP:conf/icml/IoffeS15}. 
\( Pool(\cdot) \) denotes a pooling layer that averages the input on the dimension of \( T \) in order to obtain the average representation of a sentence.
\( ReLU(\cdot) \) denotes the Rectified Linear Unit (ReLU) \cite{DBLP:conf/icml/NairH10}.
\( {C}_\ast(\cdot) \) denotes the \( 1\times1 \) convolution layer, where \( {C}_{G1}(\cdot) \) reduces the tensor’s dimension of \( E \) by a ratio of \( r \) and \( {C}_{G2}(\cdot) \) increases the dimension by the same ratio (We generally set \( r=4 \)).
Mathematically, one-dimensional convolution is equivalent to matrix multiplication.
The output of the first convolution layer and the kernel of the second convolution layer can be considered as the Q and K matrices separately in the attention calculation.
And the final result is the equivalent of an attention map to help the model dynamically select the desired information aggregation.

\begin{figure}
\centering
\includegraphics[width=0.31\textwidth]{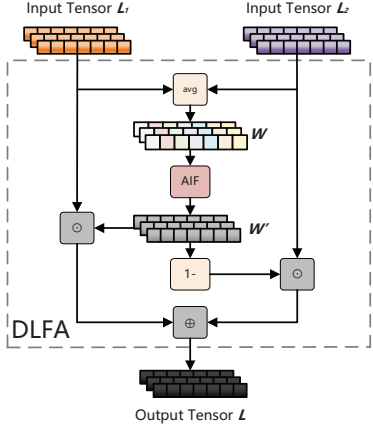}
\caption{The architecture of the proposed double layers feature aggregation (DLFA) module. The DLFA fuses the information from one selected lower transformer layer of mBERT and that from the last transformer layer.} \label{fig3}
\centering
\end{figure}

As illustrated above, the global branch first feeds the feature through a pooling layer, which is expected to capture the sentence meaning, and then get the global attention map $W_{Global}$. 
The corresponding local branch directly utilizes the convolution structure to obtain a local attention map $W_{Local}$ that focuses on each token, shown as the right branch in Fig.~\ref{fig2}. 
Compared with the branch of \( {W}_{{Global}} \), the local branch doesn’t average the input on the dimension of \( T \). 
The generated weight representing local feature \( {W}_{{Local}} \) is as follows:
 
\begin{equation}
{W}_{{Local}}=\mathcal{B}({C}_{L2}(ReLU(\mathcal{B}({C}_{L1}({W})))))
\end{equation}

\noindent where \( {C}_{L1}(\cdot) \) , \( {C}_{L2}(\cdot) \) are the functionally equivalent \( 1\times1 \) convolution layers with the convolution layers above.

At last, two branches are combined to produce the final module output \( \ {W}' \) as follows:

\vspace{-1mm}
\begin{equation}
{W}'={W}\odot Sigmoid({W}_{{Global}}\oplus{W}_{{Local}})=AIF({W})
\end{equation}

\noindent where \( \oplus \) denotes the broadcasting addition and \( \odot \) denotes the element-wise multiplication.

\subsection{Double Layers Feature Aggregation}
The double layers feature aggregation (DLFA) module is used to combine the information from the last transformer layer (12$^{th}$ layer) and from one of the lower transformers’ output to achieve better results on downstream tasks. 
As Fig.~\ref{fig3} shows,\( \ {L}_\mathbf{1} \), \( {L}_\mathbf{2} \) denote two different output features from different transformer layers (In this paper we set \( {L}_\mathbf{2} \) as the output from the 12$^{th}$ transformer layer).
And the final output \( L \) used for downstream tasks is shown as follows:

\begin{equation}
{L}={L}_\mathbf{1}\odot AIF(\frac{{L}_\mathbf{1}\oplus{L}_\mathbf{2}}{\mathbf{2}})+{L}_\mathbf{2}\odot(\mathbf{1}-AIF(\frac{{L}_\mathbf{1}\oplus{L}_\mathbf{2}}{\mathbf{2}}))
\end{equation}

\noindent where we subtract the AIF’s output weight feature from the unit tensor \( \mathbf{1} \) and element-wisely multiply the result with \( {L}_\mathbf{2} \) in order to keep the scale of the final output \( {L} \) consistent with the scale of the original transformer layer's output.

\section{Experiments}\label{sect3}
The experiments are based on the zero-shot cross-lingual transfer trials in the XTREME benchmark \cite{DBLP:conf/icml/HuRSNFJ20}.
We set English (abbr. en) as the source language, while the other languages are set as the target languages, which means we only have English training data to fine-tune mBERT for all tasks.
In this section, we introduce the datasets, the baseline system, the evaluation metrics, as well as the training details.

\begin{table*}
\caption{Characteristics of the datasets for the zero-shot transfer setting. The number of target languages (include en) for each task is shown in the $|Lang|$ column.}\label{tab1}
\begin{center}
\begin{tabular}{l|l|l|l|l|l|l|l}
\Xhline{1.5pt}
{Task} & {Corpus} & {$|Train|$} & {$|Dev|$} & {$|Test|$} & {$|Lang|$} & {Metric} & {Domain} \\ 
\hline
\multirow{2}{*}{Classification} & XNLI & 392,702 & 2,490 & 5,010 & 15 & Accuracy (Acc.) & Misc. \\ 
\cline{2-8} & PAWS-X & 49,401 & 2000 & 2000 & 7 & Accuracy (Acc.) & Wiki/Quora \\
\hline
\multirow{2}{*}{Structured Prediction.} & NER & 20,000 & 10,000 & 1,000-10,000 & 40 & F1 score (F1) & Wikipedia \\ 
\cline{2-8} & POS & 21,253 & 3,974 & 47-20,436 & 33 & F1 score (F1) & Misc. \\ \Xhline{1.5pt}
\end{tabular}
\end{center}

\end{table*}

\subsection{Dataset Description and Evaluation Metrics}
In order to demonstrate the effectiveness of our approach on cross-lingual transfer and learning linguistic structure information, we evaluate our method on four distinct datasets mentioned in the XTREME benchmark as follows:

\begin{itemize}

\item[1.]
XNLI: The Cross-lingual Natural Language Inference corpus \cite{DBLP:conf/emnlp/ConneauRLWBSS18} asks whether a premise sentence entails, contradicts, or is neutral toward a hypothesis sentence. Crowd-sourced English data is translated to ten other languages by professional translators \cite{DBLP:conf/icml/HuRSNFJ20} and used for evaluation, while the MultiNLI \cite{DBLP:conf/naacl/WilliamsNB18} training data is used for training. For simplicity, we use the average accuracy over all languages as the evaluation metric.
\item[2.]
PAWS-X: The Cross-lingual Paraphrase Adversaries from Word Scrambling \cite{DBLP:conf/emnlp/YangZTB19} dataset requires to determine whether two sentences are paraphrases. A subset of the PAWS dev and test sets \cite{DBLP:conf/naacl/ZhangBH19} is translated to six other languages by professional translators and is used for evaluation, while the PAWS training set is used for training. We also use the average accuracy over all languages as the evaluation metric.
\item[3.]
NER: For NER, we use the Wikiann \cite{DBLP:conf/acl/PanZMNKJ17} dataset. Named entities in Wikipedia were automatically annotated with LOC, PER, and ORG tags in IOB2 format using a combination of knowledge base properties, cross-lingual and anchor links, self-training, and data selection. We use the balanced train, dev, and test splits from Rahimi et al. \cite{DBLP:conf/acl/RahimiLC19}. For simplicity, we use the average F1 score over all languages as the evaluation metric.
\item[4.]
POS: We use POS tagging data from the Universal Dependencies v2.5 \cite{DBLP:conf/lrec/NivreMGHMPSTZ20} treebanks, which cover 90 languages. Each word is assigned one of 17 universal POS tags. We use the English training data for training and evaluate on the test sets of the target languages. We use the average F1 score over all languages as the evaluation metric.
\end{itemize}

Table~\ref{tab1} shows the details of datasets.
For tasks that have training and dev sets in other languages, we only report the number of English samples.

\subsection{Baseline Model}
The XTREME benchmark uses mBERT as the pre-trained multilingual language model and takes the output of the last transformer layer of mBERT as the context representation into downstream modules from the Transformers Library \cite{DBLP:journals/corr/abs-1910-03771}, which are linear classifiers corresponding to different downstream tasks. 
The whole model is first fine-tuned on English annotated data, and then tested on the unannotated data in other target languages including en to obtain evaluation results. 
We use the results of zero-shot transfer experiments in XTREME as the baseline.

\subsection{Experiment Settings of the Proposed Method}
We use mBERT as the pre-trained model for downstream tasks with the same configuration as in the XTREME benchmark. 
The difference is that the DLFA module is applied behind mBERT to obtain the fused features which are subsequently fed into the backend classifiers, as shown in Fig~\ref{fig1}. 
All the settings are consistent with that in the XTREME benchmark, and the parameters in the DLFA are Kaiming initialized \cite{DBLP:conf/iccv/HeZRS15}. 
The details of hyper-parameters for fine-tuning on mBERT are described as follows \cite{DBLP:conf/icml/HuRSNFJ20}:

We use the cased version of mBERT, which has 12 layers, 768 hidden units per layer, 12 attention heads, a 110k shared WordPiece vocabulary, and 110M parameters \cite{DBLP:conf/nips/VaswaniSPUJGKP17}. 
The model was trained using Wikipedia data in all 104 languages, oversampling low-resource languages with an exponential smoothing factor of 0.7. 
We generally fine-tune the model for five epochs, with a training batch size of 32 and a learning rate of 2e-5. 
The AdamW is adopted as the optimizer.
To construct the system conveniently, we use the Transformers Library.

\section{Results and Discussion}\label{sect4}

The results of zero-shot transfer trials for the four evaluation tasks are summarized in Table~\ref{tab2}. 
``en'' is the source language in all experiments. 
The scores in the table denote the average performance of zero-shot transfer in all target languages (i.e., within English results).
The best results of aggregation models in each task outperform the baseline by 1 to 3 absolute percentage points.
Our aggregation models achieve performance improvements on all four downstream tasks.
However, the best performances of these four tasks are obtained with different fusion layers. 
One reasonable explanation may be that these four tasks focus on different aspects of language structure learning ability, resulting in the fusion of different layers.

Wang et al. reckon that the ability to extract good semantic and structural features is a crucial reason for the model’s cross-lingual effectiveness \cite{DBLP:conf/iclr/KWMR20}. 
It is generally accepted that there exist strong similarities between two languages if they belong to the same language family \cite{DBLP:conf/acl/DongWHYW15,DBLP:journals/tacl/JohnsonSLKWCTVW17,DBLP:conf/emnlp/CotterellH17}. 
We use the ``enf'' to represent languages that belong to the same language family as English, and the ``noenf'' to represent languages that belong to the different language families from English. 
We consider that the results on the subset ``enf'' focus more on the language structure, while the results on the subset ``noenf'' focus more on the cross-lingual transfer.

\begin{table}
\caption{All results of zero-shot cross-lingual transfer trials for 4 tasks. ``D\_x'' means the system with the DLFA module that fuses the last and the x$^{th}$ transformer layers’ output.}\label{tab2}
\begin{center}
\begin{tabular}{l|c|c|c|c}
\Xhline{1.5pt}
Task     & XNLI           & PAWS-X         & NER            & POS            \\ \hline
Model$\backslash$Metrics    & Acc (\%)  & Acc (\%)  & F1 (\%)  & F1 (\%)  \\ \hline
baseline & 65.40          & 81.94          & 62.17          & 70.28          \\ \hline
D\_11    & \textbf{66.91} & 83.04          & 62.27          & 71.53          \\ \hline
D\_10    & 66.55          & 84.33 & 62.76          & \textbf{71.81} \\ \hline
D\_9     & 66.57          & 84.24          & 62.43          & 71.58          \\ \hline
D\_8     &66.90 & 82.91          & \textbf{63.34} & 71.36          \\ \hline
D\_7     & 66.20          & 83.48          & 62.63          & 71.52          \\ \hline
D\_6     & 66.75          & \textbf{84.37} & 61.84          & 71.29          \\ \hline
D\_5     & 65.42          & 82.44          & 61.66          & 71.08          \\ \hline
D\_4     & 66.14          & 82.75          & 62.28          & 71.26          \\ \hline
D\_3     & 66.00          & 83.81          & 61.88          & 71.21          \\ \hline
D\_2     & 66.15          & 83.04          & 61.34          & 71.38          \\ \hline
D\_1     & 65.85          & 82.50          & 61.73          & 71.18          \\ \Xhline{1.5pt}
\end{tabular}
\end{center}
\end{table}

\begin{table}

\begin{center}
\caption{Results of Classification tasks. The scores in the ``avg\_enf'' column denote the average performance of the subset ``enf'' (involving ``en, de''). The scores in the ``avg\_noenf'' column denote the average performance of the subset ``noenf''.}\label{tab3}
\begin{tabular}{l|c|c|l|c|c}
\Xhline{1.5pt}
\multicolumn{3}{c|}{XNLI}       & \multicolumn{3}{c}{PAWS-X}      \\ \hline
model    & avg\_enf & avg\_noenf & model    & avg\_enf & avg\_noenf \\ \hline
baseline & 75.40    & 63.86      & baseline & 89.85    & 78.80      \\ \hline
D\_11    & \textbf{77.34}  & 65.31      & D\_10    & \textbf{90.30} & 81.94      \\ \hline
D\_8     & 77.11    & \textbf{65.34}  & D\_6     & 90.00    & \textbf{82.12}  \\ \Xhline{1.5pt}
\end{tabular}
\end{center}
\end{table}

Two tasks (XNLI and PAWS-X) are selected to analyze the performance difference on the subsets ``enf'' and ``noenf''. 
As shown in Table~\ref{tab2}, there are two aggregation models, D\_11 and D\_8, with similar good results for XNLI task. 
For PAWS-X, there also exist two aggregation models, D\_10 and D\_6, that behave similarly.
We process the results of the two Classification tasks, and present the scores in Table~\ref{tab3}.
As shown in the bottom two rows of Table~\ref{tab3}, the best results of ``avg\_enf'' are obtained with systems using the information of upper layers, while the best results of ``avg\_noenf'' are obtained with systems using the information of lower layers. 
This phenomenon indicates that the information of language structure lies on the upper layers while the lower layers of mBERT are more flexible for cross-lingual transfer.

\begin{table}
\caption{Several languages' results on PAWS-X(Acc.)}\label{tab31}
\begin{center}
\begin{tabular}{l|c|c|c|c|c|c}
\Xhline{1.5pt}
Model$\backslash$Lang 	&en	&de		&fr &es		&ko	&zh        \\  \hline
mBERT 	&94.0	&85.7	&87.4	&87.0		&69.6	&77.0      \\ \hline
D\_10 	&\textbf{94.1}	&\textbf{86.5} &\textbf{88.3}	&88.79			&75.8	&80.3 \\ \hline
D\_6 	&93.9	&86.0		&87.8 &\textbf{89.09}		&\textbf{76.1}	&\textbf{81.2} \\ \Xhline{1.5pt}
\end{tabular}
\end{center}
\end{table}

Here are some specific results in PAWS-X in Table~\ref{tab31}. Scores on en, de and fr are higher in D\_10 while scores on es, ko, zh are higher in D\_6.
Coincidentally, en, de and fr are in the same language family Germanic, while es, ko and zh aren't in the Germanic family. 
This phenomenon can exactly verify our consistent conclusion that lower layers provide more cross-lingual information while upper layers provide more language structure information.
Better scores on languages not belonging to the English language family are obtained through lower-layers fusion, while upper-layers fusion gets better scores on languages within the English language family. 
The same situation also happens in the other three tasks.
So the average score embodies both cross-language ability and linguistic structural capability of the model, where the $6^{th}$ layer and the $11^{th}$ layer achieve the optimal dynamic equilibrium in Table~\ref{tab2} for PAWS-X and XNLI separately.

As shown in the tables above, the $8^{th}$ layer is the most useful for NER and the $10^{th}$ layer is the most useful for POS.
Each layer contributes differently according to different tasks.
It is necessary to consider both task-specific linguistic structures, and the generalization ability across languages when evaluated on various cross-lingual downstream tasks.
The empirical study on several datasets helps to prove the effectiveness of our proposed methods on four tasks.  
A step towards exploring the interpretability of mBERT is taken empirically.
More analyses on our method and mBERT will be conducted in the next section.

\section{Analysis Experiments}\label{sect5}
In this part, firstly, an ablation experiment is carried out to demonstrate the effectiveness of the AIF module.
Then, a toy experiment is conducted to prove our previous inference that the lower layers of mBERT are more flexible for cross-lingual transfer.
At last, the proposed method is extended on a more powerful model to prove the extensibility of our method and our conclusion.

\begin{figure}
\centering

\centering
\begin{minipage}[t]{0.7\linewidth}
\includegraphics[width=\textwidth]{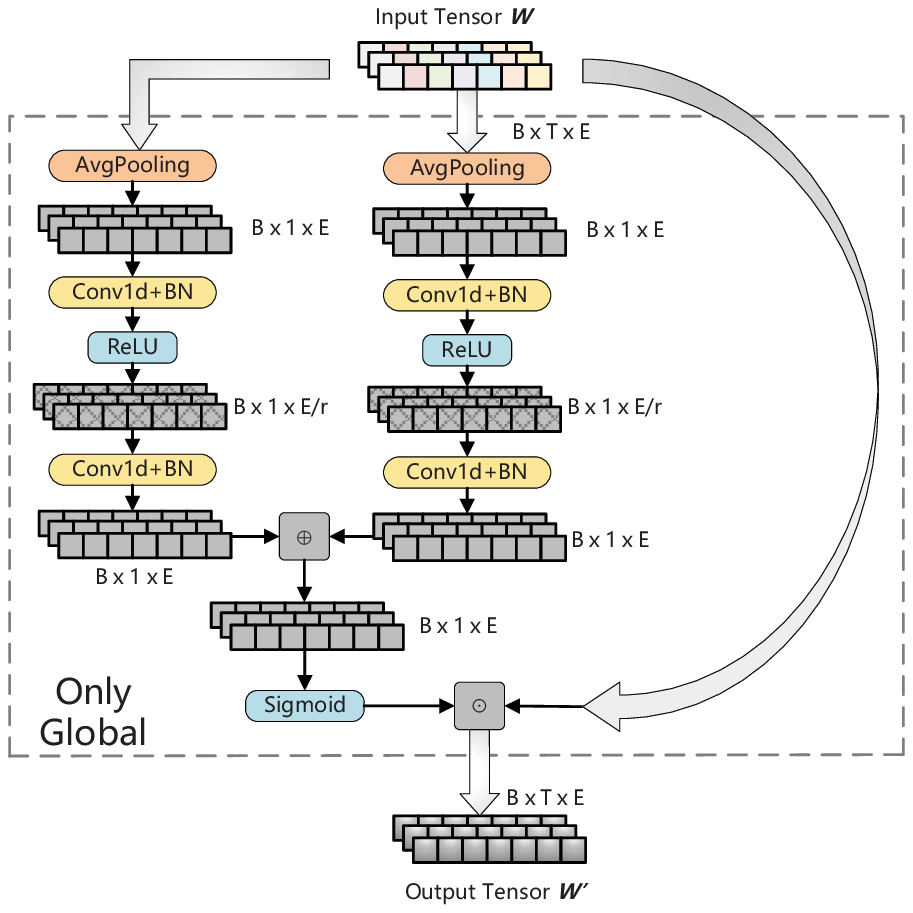}
\centering
\centerline{(a)}\medskip
\centering
\end{minipage}
\\

\centering
\begin{minipage}[t]{0.7\linewidth}
\includegraphics[width=\textwidth]{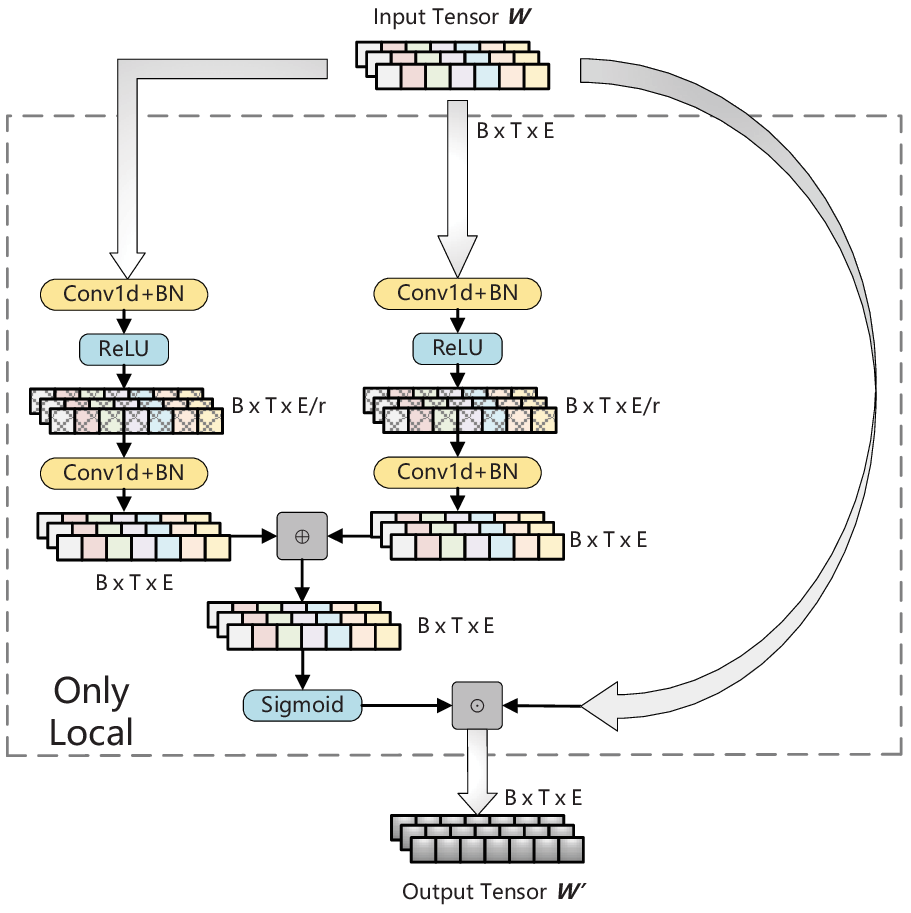}
\centering
\centerline{(b)}\medskip
\centering
\end{minipage}
\centering
\caption{(a) shows the architecture for the ablation study on the impact of only using global context. (b) shows the architecture for the ablation study on the impact of only using local context.}\label{fig4}
\vspace{-1mm}
\end{figure}

\subsection{The Ablation Study of the AIF Module}
As Fig.~\ref{fig4} shows, the aggregation architectures are redesigned to use only global or local information. 
For AIF\_global shown in Fig.~\ref{fig4}(a), we replace the local branch in the original AIF with the global branch, so the modified module will be only able to grab global information. 
Similarly, for AIF\_local represented in Fig.~\ref{fig4}(b), the global branch is replaced by the local branch, so the modified module will only grab local information.
Other experimental settings have been remained the same.
By comparing the experimental results with that of the baseline and the original AIF module, the influence of local and global information on different downstream tasks can be analyzed.

\begin{table*}
\caption{Results of the ablation study on the impact of contextual aggregation ways. The suffix ``G'' denotes only using global information and the suffix ``L'' denotes only using local information. Scores in column ``Accuracy'' or ``F1'' represent the average performance on all target languages including en.}\label{tab4}
\begin{center}
\begin{tabular}{l|c|l|c|l|c|l|c}
\Xhline{1.5pt}
\multicolumn{2}{c|}{XNLI}           & \multicolumn{2}{c|}{PAWS-X}              & \multicolumn{2}{c|}{NER}           & \multicolumn{2}{c}{POS}           \\ \hline
Model    & Accuracy                  & Model    & \multicolumn{1}{c|}{Accuracy} & Model    & \multicolumn{1}{c|}{F1} & Model    & \multicolumn{1}{c}{F1} \\ \hline
Baseline & \multicolumn{1}{c|}{65.4030} & Baseline & 81.9371                 & Baseline & 62.1700                 & Baseline & 70.2758                 \\ \hline
D\_11\_G & 66.5216                   & D\_10\_G & 83.0987                       & D\_10\_G & 62.3390                 & D\_11\_G & 71.2763                 \\ \hline
D\_11\_L & 66.3194                   & D\_10\_L & 83.4131                       & D\_10\_L & 62.1198                 & D\_11\_L & 70.7511                 \\ \hline
D\_11    & \textbf{66.9115}        & D\_10    & 84.3279                       & D\_10    & 62.7575                 & D\_11    & 71.5339                 \\ \hline
D\_10\_G & 66.4378                   & D\_6\_G  & 83.9848                       & D\_8\_G  & 62.4534                 & D\_10\_G & 71.0486                 \\ \hline
D\_10\_L & 66.2927                   & D\_6\_L  & 84.1707                       & D\_8\_L  & 62.4181                 & D\_10\_L & 70.9255                 \\ \hline
D\_10    & 66.5496                   & D\_6     & \textbf{84.3708}              & D\_8     & \textbf{63.3425}        & D\_10    & \textbf{71.8094}         \\ \Xhline{1.5pt}
\end{tabular}
\end{center}
\end{table*}

The results are displayed in Table~\ref{tab4}. 
Almost all of the models improve the performance compared to the baseline, which means both branches have positive effects.
Local information plays a more important role in the PAWS-X task,  while global information is more important in the other three tasks.
For all tasks, the original AIF module that grabs both information has the best performance. 
We can draw the conclusion that for different downstream tasks, some may focus more on local information, while others focus more on global information.
Through the AIF module, the global and local information of the initial feature integration (the average of tensors from two transformer layers) can be effectively extracted and well contextual aggregated, so as to improve the performance of the final downstream tasks.

Additionally, in order to eliminate the impact of the AIF module on classifying, we conduct a supplementary experiment D\_12 and get the results (PAWS-X: 82.01, POS: 70.34, NER: 61.91) about the same as the baseline, which may dispel the relevant concern.

\subsection{Cosine Similarity Experiment}

This experiment is an attempt to verify our findings in Sect~\ref{sect4} that the cross-lingual ability of the intermediate-level fusion model is better than that of the upper-level fusion model in XNLI and PAWS-X. 
Lample et al. use the cosine similarity and the L2 distance between word translation pairs to measure cross-lingual performance \cite{DBLP:conf/nips/ConneauL19}.
An ideal model with good cross-lingual transfer ability is language independent.
In other words, the output representations of the model should be similar when the input sentences have the same semantical meaning even if written in different languages. 
In the following toy experiment, we feed sentences with the same semantics in form of different languages (20 randomly selected parallel sentence sets from the XNLI test data, including en and 14 other languages) into different models, then extract the embeddings of these sentences. 
The average cosine similarities between the embeddings of English and that of other languages are calculated to evaluate the model's cross-lingual ability.

\begin{table}
\caption{Results of cosine similarity experiment on XNLI and PAWS-X. }\label{tab5}
\begin{center}
\begin{tabular}{l|c|l|c}
\Xhline{1.5pt}
\multicolumn{2}{c|}{XNLI}    & \multicolumn{2}{c}{PAWS-X}   \\ \hline
Model & Avg C.S. & Model & Avg C.S. \\ \hline
D\_11 & 0.5689                & D\_10 & 0.8548                \\ \hline
D\_8  & \textbf{0.6822}       & D\_6  & \textbf{0.9461}        \\ \Xhline{1.5pt}
\end{tabular}
\end{center}
\end{table}

The average cosine similarity scores with different fusion strategies are shown in Table~\ref{tab5}.
``Avg C.S.'' means the average cosine similarity between output embeddings representing information of sentences in 14 other languages and that in English.
Comparing the results in Table~\ref{tab5} and Table~\ref{tab3}, the average cosine similarity has a positive correlation with the subset “noenf” accuracy. 
The better cross-lingual ability the model has, the higher average cosine similarity between the parallel sentences is. 
This toy experiment provides us with a simple model selection method. 
For XNLI and PAWS-X tasks, the lower layers may have better cross-lingual ability than the upper layers.

\subsection{Replicate DLFA Method on XLM-R}
In order to explore the extensibility of our method and our conclusion, we replicate the proposed DLFA module on a more powerful model XLM-R base \cite{DBLP:conf/acl/ConneauKGCWGGOZ20}.
Several results are shown in Table~\ref{tab6}.
XLM-R base is also a 12 layer cross-language model, and the results prove that our method can also work on other models, showing great generalization ability across different models.
It is worth noting that the most important layers in XLM-R-base for different tasks are just the same with mBERT.
Since the basic pre-training methods of these two models are the same, the common characters observed are worth further exploration.

\begin{table}
\caption{Results of XLM-R-base.}\label{tab6}
\begin{center}
\begin{tabular}{l|c|c|c}
\Xhline{1.5pt}
Model$\backslash$Task       & PAWS-X(Acc.)  & POS(F1)  & NER(F1)        \\  \hline
XLM-R-base  &82.67	&71.44	&61.09      \\ \hline
The proposed method  &\textbf{85.10}  &\textbf{73.09}	&\textbf{62.85} \\ \hline
(best layer) &6 &8 &10 \\ \Xhline{1.5pt}
\end{tabular}
\end{center}
\end{table}

\section{Conclusion}\label{sect6}
Different layers contain diverse cross-language information and language structure information, which are complementary to each other for downstream tasks. 
Only using the last layer of mBERT in most downstream tasks does not make full use of diverse information contained in different mBERT’s layers.
In this paper, we propose the DLFA module to integrate the information of the last transformer layer with other layers of mBERT, and apply the optimal aggregation models to different tasks.
We conduct a series of experiments and explore the interpretability of mBERT empirically.
Through the results, we prove the effectiveness of the proposed method and come to a consistent conclusion about the distribution of cross-lingual capability and language structured ability of mBERT. In the future work, we will pay more attention to the rich information contained in different layers of mBERT, and apply them to more downstream tasks.


\end{document}